\documentclass[11pt,a4paper]{article}
\usepackage[hyperref]{emnlp2020}
\usepackage{times}
\usepackage{latexsym}
\usepackage{epsfig}
\usepackage{graphicx}
\usepackage{amsmath}
\usepackage{amssymb}
\usepackage{enumitem}
\usepackage{tikz}
\usepackage{pgfplots}
\pgfplotsset{compat=1.14}  
\usepgfplotslibrary{colorbrewer}
\usepackage{caption} 
\usepackage{booktabs}
\usepackage{algorithm} 
\usepackage{algpseudocode}

\usepackage{microtype}

\aclfinalcopy 
\def\aclpaperid{995} 

\title{Sub-Instruction Aware Vision-and-Language Navigation}

\author{Yicong Hong$^{\ast 1}$, Cristian Rodriguez-Opazo$^{\ast 1}$, Qi Wu$^{2}$, Stephen Gould$^{1}$ \\
  $^{1}$The Australian National University, $^{2}$University of Adelaide \\
  $^{1,2}$Australian Centre for Robotic Vision \\
  \small \texttt{\{yicong.hong, cristian.rodriguez, stephen.gould\}@anu.edu.au} \\ 
  \small \texttt{qi.wu01@adelaide.edu.au}} 

\date{}

\begin{document}
\maketitle
\begin{abstract}
Vision-and-language navigation requires an agent to navigate through a real 3D environment following natural language instructions. Despite significant advances, few previous works are able to fully utilize the strong correspondence between the visual and textual sequences. Meanwhile, due to the lack of intermediate supervision, the agent's performance at following each part of the instruction cannot be assessed during navigation. In this work, we focus on the granularity of the visual and language sequences as well as the traceability of agents through the completion of an instruction. We provide agents with fine-grained annotations during training and find that they are able to follow the instruction better and have a higher chance of reaching the target at test time. We enrich the benchmark dataset Room-to-Room (R2R) with sub-instructions and their corresponding paths. To make use of this data, we propose effective sub-instruction attention and shifting modules that select and attend to a single sub-instruction at each time-step. We implement our sub-instruction modules in four state-of-the-art agents, compare with their baseline models, and show that our proposed method improves the performance of all four agents.

We release the Fine-Grained R2R dataset (FGR2R) and the code at \texttt{\url{https://github.com/YicongHong/Fine-Grained-R2R}}. {\let\thefootnote\relax\footnote{{* Authors contributed equally}}}
\end{abstract}

\section{Introduction}
\label{sec:intro}

\begin{figure}[t]
   \centering
   \includegraphics[width=0.48\textwidth]{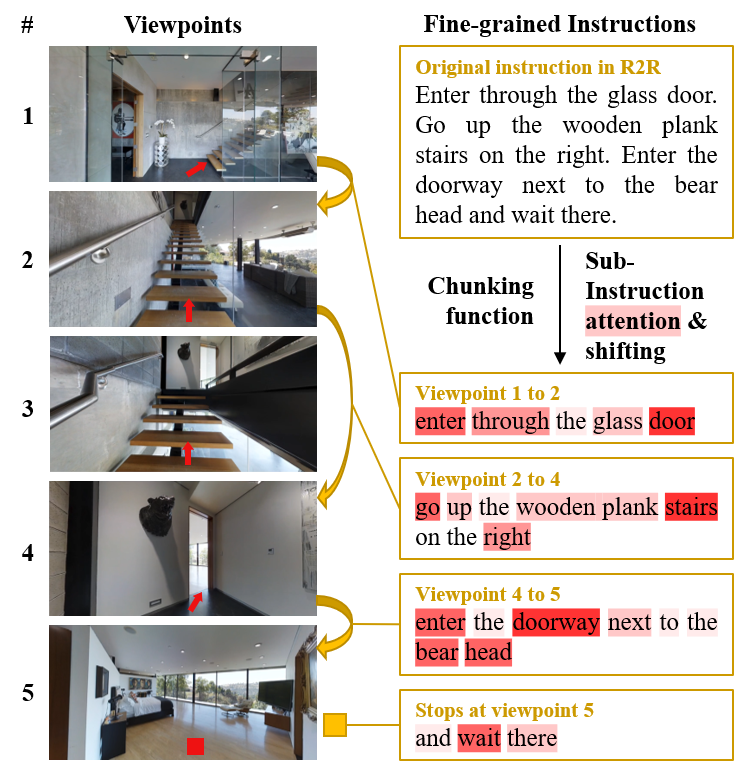}
   \caption{Visual navigation with sub-instruction and sub-path pairs. We enrich the R2R dataset by providing fine-grained matching between sub-instructions and viewpoints along the ground-truth path.}
   \label{fig:intro}
\end{figure}

Creating an agent that can navigate through an unknown environment following natural language instructions has been a dream of human-beings for many years. Such an agent needs to possess the ability to perceive its environment, understand the instructions and learn the relationship between these two streams of information. Recently, \citeauthor{anderson2018vision} (\citeyear{anderson2018vision}) proposed the vision-and-language navigation (VLN) task that formalized such requirements through an evaluation of an agent’s ability to follow natural language instructions in photo-realistic environments.

Despite the significant progress made by recent approaches, there is little evidence that agents learn the correspondence between observations and instructions. \citeauthor{hu2019you} (\citeyear{hu2019you}) found that a modified self-monitoring agent \cite{ma2019self}, could achieve similar performance with (success rate 40.5\%) and without (success rate 39.7\%) visual information. Among other reasons, such as dataset bias, the result suggests that this agent gains little from having the two streams of information.

We argue that one of the main reasons behind this is that current methods are not adequately teaching agents the relationship between perception---things that the robot is observing---and parts of the instructions. Since datasets do not provide such information agents can only use the ground-truth trajectory as a learning signal.  Moreover, given the lack of fine-grained annotation, current methods cannot evaluate the (perceptual or linguistic) grounding process at each step as there is no ground truth signal to indicate which part of the instruction has been completed.

To address this problem, we enhance the R2R dataset \cite{anderson2018bottom} to acquire intermediate supervision for the agents, providing a fine-grained matching between sub-instructions and the agent's visual perception, as illustrated in Figure~\ref{fig:intro}, to produce our Fine-Grained Room-to-Room dataset (FGR2R). We argue that the granularity of the navigation task should be at the level of these sub-instructions, rather than attempting to ground a specific part of the original long and complex instruction without any direct supervision or measure navigation progress at word level.

Our work aims to make the navigation process traceable and encourage the agent to run precisely on the described path rather than just focusing on reaching the target. We hypothesize that the agent should reach the target with higher success rate by following a detailed instruction with richer information, and in practice, the agent could complete some additional tasks on its way to the target. 

In light of this, we propose a novel sub-instruction attention mechanism to better learn the correspondence between visual features and language features. Our agent first segments the long and complicated instruction into short and easier-to-understand sub-instructions using a heuristic method based on the grammatical relations provided by the Stanford NLP Parser \cite{qi2019universal}. Moreover, we propose a shifting module that infers whether the current sub-instruction has been completed. Hence, only one sub-instruction is available to the agent at each time step for textual grounding. These modules can be easily applied to previous VLN models.

We conduct experiments to compare the performance of four state-of-the-art agents to evaluate with or without our sub-instruction module, for agents based on imitation learning \cite{anderson2018vision, fried2018speaker, ma2019self} and reinforcement learning \cite{tan2019learning}. Analyzing the results we find that the intermediate supervision and our proposed modules help the agents to better follow the instructions. Furthermore, we demonstrate the traceability of the navigation process through qualitative and quantitative analysis.

\section{Related Work}
\label{sec:related_work}

\noindent \textbf{Visual and textual grounding.} Visual grounding aims to infer the relationship between a text description and a spatial or temporal region in an image or video, respectively. It is an essential component for a variety of tasks in vision-and-language research such as visual question answering (VQA) \cite{schwartz2017high, anderson2018bottom, hudson2019learning}, image captioning \cite{xu2015show, anderson2018bottom, cornia2019show, ma2020learning}, video understanding \cite{gao2017tall, ma2018attend, opazo2019proposal} and phrase localization \cite{engilberge2018finding,yu2018mattnet}. In the case of vision-and-language navigation, at each navigational step, the agent attends to the relevant part of the instruction according to visual clues to direct the future action. Meanwhile, the agent attends the visual inputs at different directions as described by text to perceive the environment \cite{fried2018speaker, ma2019self}.

\medskip

\noindent \textbf{Vision and language navigation.} 
\citeauthor{anderson2018vision} (\citeyear{anderson2018vision}) formalized the vision-and-language navigation task in a photo-realistic environment, and proposed a benchmark Room-to-Room (R2R) dataset and a sequence-to-sequence agent as a baseline model. Other datasets in real environments, such as R4R \cite{jain2019stay}, which is an extended version of R2R with longer instruction-path pairs, and Touchdown \cite{chen2019touchdown} for navigation on streets have also been proposed for study. 

Researchers have addressed the R2R task through a great variety of approaches. \citeauthor{wang2018look} (\citeyear{wang2018look}) propose a look-ahead model that combines model-based and model-free reinforcement learning, predicts the agent's next state and reward during navigation. \citeauthor{fried2018speaker} (\citeyear{fried2018speaker}) proposed the Speaker-Follower model which generates augmented samples for training and makes use of the panoramic action space to ground and navigate efficiently. Later, \citeauthor {ma2019self} (\citeyear{ma2019self}) introduced the Self-Monitoring agent which includes a vision and language co-grounding network and a progress monitor. The progress monitor estimates a normalized distance to the target and guides the transition of the textual attention. \citeauthor{wang2019reinforced} (\citeyear{wang2019reinforced}) applied the REINFORCE algorithm \cite{williams1992simple} to improve the agent's generalizability and proposed a Self-Supervised Imitation Learning (SIL) method to facilitate lifelong learning in a new environment. The Back Translation agent \cite{tan2019learning} applied the A2C algorithm \cite{mnih2016asynchronous}  and made use of a speaker module with environmental dropout for data augmentation. \citeauthor{landi2019embodied} (\citeyear{landi2019embodied}) applied dynamic convolutional filters for image feature extraction for low-level grounding of visual inputs and \citeauthor{hu2019you} (\citeyear{hu2019you}) grounded multiple modalities using a mixture-of-experts approach and applied joint training strategy. Besides, the Regretful agent \cite{ma2019regretful} and the Tactical Rewind agent \cite{ke2019tactical} are models which focus on path-scoring and backtracking methods. Very recently, \citeauthor{zhu2020vision} (\citeyear{zhu2020vision}) introduces multiple auxiliary losses in training to help exploring the semantic meaning of visual features, \citeauthor{huang2019transferable} (\citeyear{huang2019transferable}) and \citeauthor{hao2020towards} (\citeyear{hao2020towards}) apply pre-trained encoders to generate generic visual and textual representations for the agent.

In contrast to all previously mentioned methods that ground the complete instruction, we propose to divide the instruction into meaningful semantic sub-instructions, and teach the agent to complete each one at a time before reading the next sub-instruction. In that spirit, our method is similar to the image captioning work by \citeauthor{cornia2019show} (\citeyear{cornia2019show}). They design a shifting gate over the image regions to control the visual features that feed into each time step of the caption module. We differ from their work in the modality that is attended. Our method works in the language domain, and the shifting depends only on local context rather than looking over all the sub-instructions. BabyWalk \cite{zhu2020babywalk} is a concurrent work to ours, it uses sub-instructions for curriculum learning which trains the agent to complete shorter navigation tasks before trying to solve the longer ones. Comparing the sub-instruction and sub-path pairs in FGR2R and BabyWalk, BabyWalk aligns the textual and visual sequences by solving a dynamic programming problem, whereas FGR2R employs human annotation, which is more fine-grained and accurate.

\section{Sub-instruction Aware VLN}
\label{sec:method}
In this section, we first introduce the VLN problem and the general architecture of the agent. Then, we discuss about the proposed chunking function for producing the sub-instructions and the novel sub-instruction module for enabling sub-instruction attention and transition.

The VLN task requires the agent to navigate through a real environment to a target location following a natural language instruction. Formally, an instruction $\boldsymbol{w}$ is a sequence of words $\langle w_{1},w_{2},\ldots,w_{l}\rangle$ provided to the agent at the beginning of its navigation, where $w_{i}$ denotes the $i$-th word in the sequence. 
The environment is defined as set of viewpoints $\{p_j\}$ denoting all the navigable locations. At time step $t$, the agent at viewpoint $p_t$ receives a panoramic view $\boldsymbol{V}_{\!t}$ composed of $n$ single view images $\langle \boldsymbol{v}_{t,1},\boldsymbol{v}_{t,2},\ldots,\boldsymbol{v}_{t,n}\rangle$. Using the given instruction $\boldsymbol{w}$ and the current observation $\boldsymbol{V}_{\!t}$, the agent needs to infer an action $a_{t}$ which triggers a transition signal from $p_t$ to $p_{t+1}$. The episode ends when the agent output a $STOP$ action or the maximum number of steps allowed is reached.

\begin{figure}[t]
   \centering
   \includegraphics[width=0.47\textwidth]{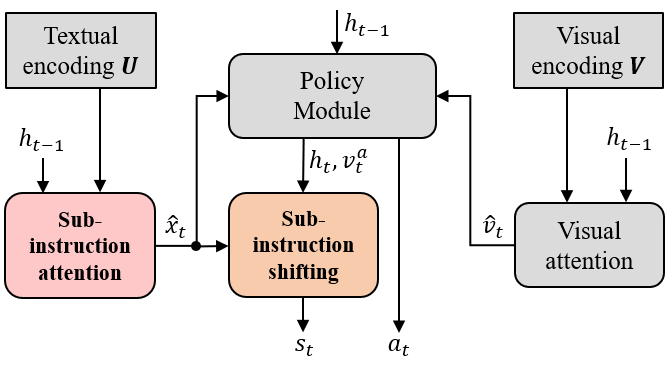}
   \caption{Our sub-instruction attention and shifting modules built into the self-monitoring agent pipeline. We replace the original textual attention module with our sub-instruction modules that select and attend a single sub-instruction at each time-step.}
   \label{fig:pipeline}
\end{figure}

\subsection{Base Agent Model}
We build our sub-instruction module based on the current state-of-the-art VLN agents, as shown in Figure \ref{fig:pipeline}. Those agents share a similar pipeline, a sequence-to-sequence architecture with textual and visual attentions. In this section, we refer to the Self-Monitoring Agent \cite{ma2019self} to present the flow of information in the network.

\medskip

\noindent \textbf{Visual and textual encoding}. Before the start of navigation, the agent first encodes the given instruction, using an LSTM with a learned embedding as $\boldsymbol{\hat{w}}_{j}{=}$ Embed$(w_{j})$ and $\boldsymbol{u}_{1},\boldsymbol{u}_{2},\ldots,\boldsymbol{u}_{l}{=}$ LSTM$(\boldsymbol{\hat{w}}_{1},\boldsymbol{\hat{w}}_{2},\ldots,\boldsymbol{\hat{w}}_{l})$, where $\boldsymbol{u}_{j}$ is the hidden state of word $w_{j}$ in the instruction.
In the case of the panoramic view, the agent encodes the images using a ResNet-152 model \cite{he2016deep} pre-trained on ImageNet \cite{ILSVRC15} for each navigable direction. A 4-dimensional vector $[\sin\psi,\cos\psi,\sin\theta,\cos\theta]$ is concatenated with the image encoding to represent the direction of visual features, where $\psi$ and $\theta$ are the heading and elevation angles, respectively.

\medskip
\noindent \textbf{Policy module and co-grounding}. We define the agent's state at time $t$ as a combination of the attended textual representation $\boldsymbol{\hat{u}}_{t}$, the attended visual representation $\boldsymbol{\hat{v}}_{t}$ and the previous selected action $\boldsymbol{a}_{t-1}$, encoded by an LSTM as
\begin{equation}
\boldsymbol{h}_{t}, \boldsymbol{m}_{t}= \text{LSTM}([\boldsymbol{\hat{v}}_{t};\boldsymbol{a}_{t-1}], (\boldsymbol{\hat{u}}_{t}, \boldsymbol{m}_{t-1})).
\label{eqn:agent_state}
\end{equation}
We refer to $\boldsymbol{h}$ and $\boldsymbol{m}$ as the agent's state and memory, respectively.

\noindent The attended textual representation is obtained by performing soft-attention over the language features $\boldsymbol{U}{=}\langle \boldsymbol{u}_{1},\boldsymbol{u}_{2},\ldots,\boldsymbol{u}_{l}\rangle$ with the agent's state at the previous time step. The attention weights over all the words are calculated as $z^{\text{text}}_{t,j}{=}(\boldsymbol{W}_{\!u}\boldsymbol{h}_{t-1})^{T}\boldsymbol{u}_{j}$ and $\boldsymbol{\alpha}_{t}{=}$ Softmax$(\boldsymbol{z}^{\text{text}}_{t})$, obtaining the attended textual representation by $\boldsymbol{\hat{u}}_{t}{=}{\boldsymbol{\alpha}_{t}}^{T}\boldsymbol{U}$.
Similarly, we perform soft-attention over the single-view visual features $\boldsymbol{V_{t}}$ as $z^{\text{vis}}_{t,i}{=}(\boldsymbol{W}_{\!v}\boldsymbol{h}_{t-1})^{T}g(\boldsymbol{v}_{t,i})$ where $g(\cdot)$ is a multi-layer perceptron (MLP), and the attention weight $\boldsymbol{\beta}_{t}{=}$ Softmax$(\boldsymbol{z}^{\text{vis}}_{t})$. The attended visual representation is $\boldsymbol{\hat{v}}_{t}{=}{\boldsymbol{\beta}_{t}}^{T}\boldsymbol{V_{t}}$. The previous selected action $\boldsymbol{a}_{t-1}$ is represented by the visual features at the previously selected action direction. 
Finally, the agent decides an action by finding the visual features at a navigable direction with the highest correspondence to the attended language features $\boldsymbol{\hat{u}}$ and the agent's current state $\boldsymbol{h}_{t}$. The probability at each navigable direction is computed as:
\begin{equation}
o_{t,i}=(\boldsymbol{W}_{a}[\boldsymbol{h}_{t},\boldsymbol{\hat{u}}_{t}])^{T}g(\boldsymbol{v}_{t,i})
\label{eqn:action_prob_1}
\end{equation}
and
\begin{equation}
\boldsymbol{p}_{t}=\text{Softmax}(\boldsymbol{o}_{t})
\label{eqn:action_prob_2}
\end{equation}
where $g(\cdot)$ is the same MLP as in visual attention for feature projection. Then, the agent moves in a panoramic action space \cite{fried2018speaker}, so that it jumps directly to an adjacent viewpoint in the selected direction.

All baseline agents in our experiments are variants of this pipeline. For instance, the Speaker-Follower agent \cite{fried2018speaker} encodes the agent's state with only the previous action and the attended visual features. In the case of the Back-Translation agent \cite{tan2019learning}, it attends the language features by the agent's current state.

\medskip
\subsection{Chunking}
\label{sec:Methods:chunk_gen}

To encourage the learning of vision and language correspondences, we provide short and easier-to-learn sub-instructions to the agent at each time step. Formally, for each instruction $\boldsymbol{w}$, there exists a set of sub-instructions $\boldsymbol{X}{=}\langle \boldsymbol{x}_{1},\boldsymbol{x}_{2},...,\boldsymbol{x}_{L}\rangle$, where $\boldsymbol{x}_{i}{=}\langle w_{j}\rangle$ and $L$ is the total number of sub-instructions. The sub-instructions are ordered, mutually exclusive and cover the entire $\boldsymbol{w}$.

We propose a chunking function to break the original instruction into several sub-instructions, where each sub-instruction is an independent navigation task and usually requires the agent to perform one or two actions to complete. To achieve this automatically, we design chunking rules based on the grammatical relations between words in the instruction, where the relations are produced by the Stanford NLP Parser \cite{qi2019universal}, a pre-trained natural language analysis tool. First, we pass the entire instruction into the StanfordNLP Parser for extracting the $dependency$ and the $governor$ of each word, denoted as $\eta(w_{j})$ and $\rho(w_{j})$, respectively. Then, using the two attributes, we formulate a heuristic as shown in Algorithm~\ref{alg:chunking_function}. 

\begin{algorithm}
	\caption{Chunking Function}
\small
	\begin{algorithmic}[!bth]
	    \State Initialize empty lists $l_{conj}$, $l_{x}$, $l_{\eta}$, $l_{\boldsymbol{X}}$. Count $k=0$.
	    \State \# Find index of the word that satisfies condition (2)
		\For {$w_{j}$ in $\boldsymbol{w}$} 
		   	\If {$\eta(w_{j})$ \text{is} $\texttt{conj}$ \&\& $\rho(w_{j})$ \text{is} $1$}
        		\State Save word index $j$ into $l_{conj}$
    		\EndIf
    	\EndFor
		\For {$w_{j}$ in $\boldsymbol{w}$}
		    \State \# Condition (1)
    	   	\If {$\eta(w_{j})$ \text{is} $\texttt{root}$ \&\& ($\texttt{root}$ \text{in} $l_{\eta}$ or $\texttt{parataxis}$ \text{in} $l_{\eta}$)}
        	    \State $l_{\boldsymbol{X}}$ $\leftarrow$ \textbf{Check}$(l_{x})$
            \State \# Condition (2) 
        	\ElsIf {$k\leq\text{len}(l_{conj})-1$ \&\& $\rho(w_{j})$ is $l_{conj}[k]$}
        	    \State $l_{\boldsymbol{X}}$ $\leftarrow$ \textbf{Check}$(l_{x})$, $k=k+1$
            \State \# Condition (3)
        	\ElsIf {$\eta(w_{j})$ \text{is} $\texttt{parataxis}$ \&\& ($\texttt{root}$ \text{in} $l_{\eta}$ or $\texttt{parataxis}$ \text{in} $l_{\eta}$)}
        	    \State $l_{\boldsymbol{X}}$ $\leftarrow$ \textbf{Check}$(l_{x})$
        	\State \# Save the word into temporary chunk
        	\ElsIf {$\eta(w_{j})$ is not \texttt{punct}}
        	    \State Save $w_{j}$ into $l_{x}$, save $\eta(w_{j})$ into $l_{\eta}$
    		\EndIf
		\EndFor
	\end{algorithmic}
\label{alg:chunking_function}
\end{algorithm}

The chunking function considers words in the instruction that meet one of the following three conditions as the beginning of a new sub-instruction: (1) its dependency is \texttt{root} and all the words before belong to the previous chunk, (2) its dependency is \texttt{conj} and its governor is the previous \texttt{root}, (3) its dependency is \texttt{parataxis} and all the words before belong to the previous chunk. If any one of the three conditions is met, a \textbf{Check}$(\cdot)$ function will be performed on the temporary chunk to decide whether to save the temporary chunk into the final sub-instruction list $l_{\boldsymbol{X}}$. Here, the \textbf{Check}$(\cdot)$ function examines if the temporary chunk meets two conditions: (1) the chunk length should exceed the minimum length of two words, and (2) the temporary chunk should not only contains a single action-related phrase which is following the previous chunk or is leading the next chunk (e.g. ``\textit{go straight then ...}"), if it happens, then the temporary chunk should be appended to the previous chunk or added to the next chunk respectively.

We provide an illustrative example here. Our chunking function breaks the given instruction ``\textit{Enter through the glass door. Go up the wooden plank stairs on the right. Enter the doorway next to the bear head and wait there}." into \fbox{1} ``\textit{Enter through the glass door}", \fbox{2} ``\textit{Go up the wooden plank stair on the right}", \fbox{3} ``\textit{Enter the doorway next to the bear head}" and \fbox{4} ``\textit{And wait there}", as shown in Figure \ref{fig:intro}. In the third and the fourth sub-instructions, the words ``Enter" and ``Wait" satisfy the conditions (1) and (2), respectively. Notice that the $governor$ of conjunction word ``And" is ``Wait", so it has been assigned to the fourth sub-instruction.

\subsection{Sub-Instruction Module}
To encourage the agent to learn the correspondence between visual and language features in a sub-instruction, we modify the base agents to include a sub-instruction module, which enables the agent to focus on a particular sub-instruction at each time step, as shown in Figure \ref{fig:pipeline}. It contains two main components: the sub-instruction attention and the sub-instruction shifting module.

\medskip
\noindent \textbf{Sub-instruction attention.} The module attends the words inside the selected sub-instruction $\boldsymbol{x}_{i}$ through a soft-attention mechanism. Formally, at each time step, we calculate the distribution of weights over each word in  $\boldsymbol{x}_{i}$ as:
\begin{align}
z^{\text{text}}_{t,j}&=(\boldsymbol{W}_{\!u}\boldsymbol{h}_{t-1})^{T}\boldsymbol{x}_{i,j}, \\ \nonumber
\boldsymbol{\alpha}_{t}&=\text{Softmax}(\boldsymbol{z}^{\text{text}}_{t})
\label{eqn:chunk_attention}
\end{align}
where $\boldsymbol{h}_{t-1}$ is the previous state of the agent and $\boldsymbol{W}_{\!u}$ is the learned weights. The grounded representation of the sub-instruction is hence $\hat{\boldsymbol{x}}_{i}={\boldsymbol{\alpha}_{t}}^{T}\boldsymbol{x}_{i}$.

With the sub-instruction attention, the agent is forced to attend the most relevant part of the instruction and prevent the agent from ``getting distracted'' by the other part of the instruction that has been completed or to be completed in the further steps.

\medskip
\noindent \textbf{Sub-instruction shifting.} At each time step, the agent needs to decide whether the current sub-instruction will be completed by the next action or not. 
We enable this functionality by designing a shifting module that estimates the probability of proceeding to the next sub-instruction.

The module uses a recurrent neural architecture to encode a representation that reflect the vision and language co-grounded features:
\begin{equation}
\boldsymbol{h}^{c}_{t} = \sigma(\boldsymbol{W}_{\!c1}[\boldsymbol{W}_{\!c0}(\boldsymbol{h}_{t}),\boldsymbol{v}^{a}_{t}, \hat{\boldsymbol{x}}_{i}])\odot\text{tanh}(\boldsymbol{m}_{t})
\label{eqn:chunk_shift_1}
\end{equation}
where $\boldsymbol{h}_{t}$ and $\boldsymbol{m}_{t}$ is the agent's current state and memory, $\boldsymbol{v}^a_{t}$ is the visual feature at the selected action direction, $\sigma$ represents a sigmoid function, $\boldsymbol{W}_{\!c1}$ and $\boldsymbol{W}_{\!c0}$ are the learned weights and $\odot$ denotes the Hadamard product.

The module then computes the shifting probability from $\boldsymbol{h}^{c}_{t}$ and a one-hot encoding $\boldsymbol{e}_{t}$ of the number of sub-instructions left to be completed, as:
\begin{equation}
{p}^{s}_{t}=\sigma(\boldsymbol{W}_{\!c2}[\boldsymbol{W}_{\!c3}(\boldsymbol{e}_{t}),\boldsymbol{h}^{c}_{t}])
\label{eqn:chunk_shift_2}
\end{equation}
where $\boldsymbol{W}_{\!c2}$ and $\boldsymbol{W}_{\!c3}$ are the learned parameters. Here, $\boldsymbol{e}_{t}$ introduces a learnable prior on when to shift before viewing the scene. This prior is then modified by taking into account the visual evidence, which is essential for efficient navigation. If the shifting probability exceed a certain threshold, a shift signal ${s}_{t}{=}1$ (${s}_{t}{\in}{\{0,1\}}$) of reading the next sub-instruction will be produced. We only enable the module to do a single step uni-directional shifting, which agrees with the fact that instructions and trajectory in the R2R dataset are monotonically aligned.

\subsection{Training}

In the training stage, for each instruction $\boldsymbol{w}$, there exists a corresponding ground-truth path $\boldsymbol{p}_{g}=\langle p_{g(1)},p_{g(2)},...,p_{g(M)}\rangle$. In the case of sub-instructions, we partition the path into sub-paths, one for each sub-instruction.
 
The binary cross-entropy loss compares the estimated shifting probabilities $p^{s}_{t}$ to the target shifting signals $y^{s}_{t}$, where the target is either 1 or 0 depending on the distance between the agent's current position and the ending viewpoint of the current sub-path. In summary, the agent's parameters are learned to optimized
\begin{align}
\mathcal{L}=&-\sum_{t}y^{a}_{t}\log{p}_{t}^{a} - \\ \nonumber
&\sum_{t} y^{s}_{t}\log{p}^{s}_{t}+(1-y^{s}_{t})\log(1-{p}^{s}_{t})
\label{eqn:objective}
\end{align}
where $p^{a}_{t}$ is the predicted action, $y^{a}_{t}$ and $y^{s}_{t}$ are the ground-truth action and shifting signal respectively at time step $t$.

During training, we apply student-forcing supervision to the action to encourage exploration, but use teacher-forcing for the sub-instruction shifting \cite{williams1989learning, anderson2018vision}. In early stages of training, the ground-truth shifting signal will have a large number of zeros since the agent has a high probability of deviating from the desired path. We prevent the sub-instruction shifting module from converging to an undesirable local minimum by forcing the shifting loss to consider an equal number of randomly selected shift and do-not-shift samples in each time step.
\section{The FGR2R Dataset}
\label{sec:FGR2R}

To acquire the matching between vision and language sub-sequences, we introduce a Fine-Grained Room-to-Room (FGR2R) dataset which enriches the benchmark Room-to-Room dataset by dividing the instructions into sub-instructions and pairing each of those with their corresponding viewpoints in the path.

\noindent \textbf{Dataset collection.}
We first apply the chunking function introduced in Section \ref{sec:Methods:chunk_gen} to generate the sub-instructions automatically from the original R2R data. We demonstrate the quality of the generated sub-instructions by comparing the output sub-instructions against a manually annotated subset of 300 samples, obtaining a smoothed BLEU-4 score of $0.84$. Then, we add annotations of sub-path corresponding to each sub-instruction using the Amazon Mechanical Turk (AMT)\footnote{Amazon Mechanical Turk: https://www.mturk.com/}. We refer the readers to Appendix \ref{sec:appendix_A1} for more information about the data collection interface and the qualification process of the annotators that we designed to ensure the quality of the collected data.

\noindent \textbf{Dataset statistics.}
The original R2R possesses 21,567 navigation instructions and 7,189 paths in 91 real-world environments, where 3 or 4 different natural language instructions describe each path. The R2R data has been split for learning proposes, with 4,675 paths for training and 340 paths for seen validation in 61 scenes, 783 paths in 11 scenes for unseen validation and the remaining 1,391 paths in 18 scenes for testing\footnote{More information about R2R can be found in the Matterport3D dataset\cite{chang2017matterport3d} and the R2R dataset \cite{anderson2018vision}}. Based on the original R2R data, FGR2R divides the instructions for the training and validation set in an average of $3.6$ sub-instructions. Each sub-instruction has $7.2$ words on average. Sub-instructions are paired on average $2.4$ viewpoints, and with a minimum and maximum of $1$ and $7$ viewpoints, respectively. We refer the readers to Appendix \ref{sec:appendix_A1} for more dataset statistics.


\section{Experiments} 
\label{sec:experiments}

\subsection{Experiment Setup}
We experiment with four state-of-the-art VLN agents with and without our sub-instruction module and compare their performance on the original R2R validation unseen split. 

The agents are chosen to include the most common network architectures, training strategies and inference methods among the previous VLN agents. They include the Sequence-to-Sequence (Seq2Seq) \cite{anderson2018vision} model which does not apply panoramic action space, two visual-textual co-grounding models, the Speaker-Follower \cite{fried2018speaker} and the Self-Monitoring agent \cite{ma2019self}, as well as the Back-Translation model \cite{tan2019learning} which applies reinforcement learning. For all agents, we implement our sub-instruction module in their network based on their officially released code. For the self-monitoring agent, we remove the progress monitor since it requires the attention weight over the entire instruction for estimating the navigation progress.

\begin{figure*}[t]
  \centering
  \includegraphics[width=0.99\textwidth]{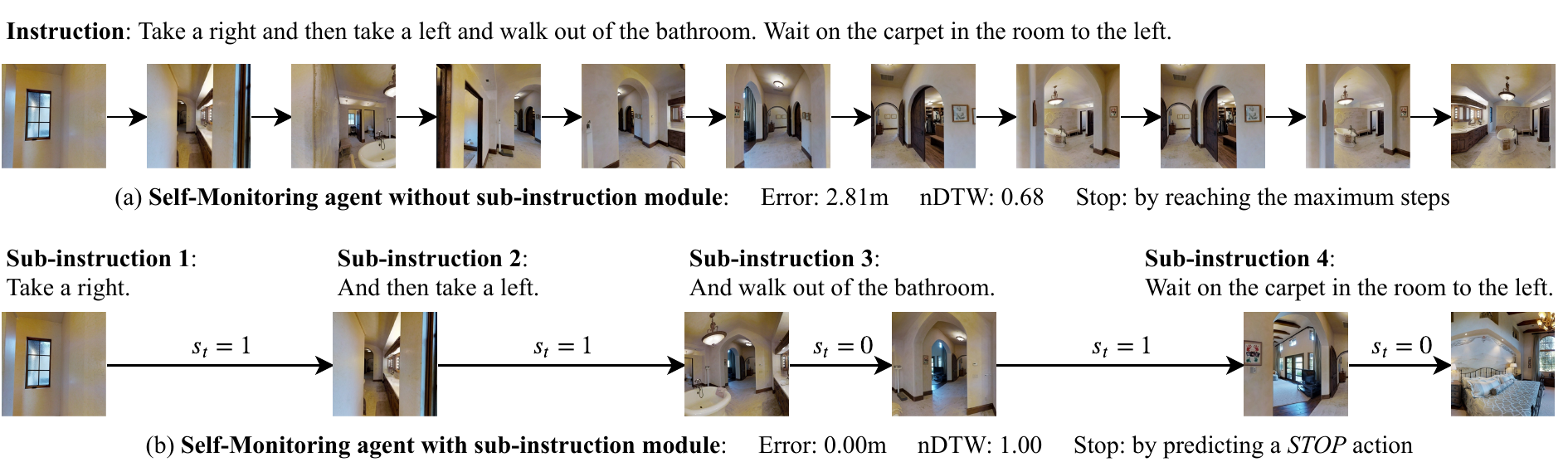}
  \caption{Qualitative comparison of a successful case without and with sub-instruction module. Without sub-instruction module, the agent fails to follow the instruction and stops next to the target by chance. With sub-instruction module, the agent navigates on the described path and eventually stops right at the target location. For panoramic visualization and more examples please refer to the supplementary material.}
  \label{fig:path_comparison}
\end{figure*}

\begin{table*}[t]
\centering
\resizebox{\textwidth}{!}{\begin{tabular}{llrrrrrr}
\toprule
& & \multicolumn{6}{c}{R2R Validation Unseen} \\
\cmidrule(r){3-8}
\# & Model & \multicolumn{1}{c}{PL $\downarrow$} & \multicolumn{1}{c}{NE $\downarrow$} & \multicolumn{1}{c}{OSR $\uparrow$} & \multicolumn{1}{c}{SR $\uparrow$} & \multicolumn{1}{c}{SPL $\uparrow$} & \multicolumn{1}{c}{nDTW $\uparrow$} \\
\cmidrule(r){1-8}

1 & Seq2Seq \cite{anderson2018vision} & \textbf{8.34} (8.71) & \textbf{7.85} (7.92) & 29.2 (\textbf{29.5}) & \textbf{22.9} (21.8) & \textbf{0.20} (0.18) & \textbf{0.58} (0.57) \\

2 & Speaker-Follower \cite{fried2018speaker} & \textbf{13.57} (16.66) & \textbf{6.66} (7.12) & \textbf{44.8} (41.1) & \textbf{34.7} (29.8) & \textbf{0.28} (0.22) & \textbf{0.59} (0.54) \\

3 & Self-Monitoring \cite{ma2019self} & \textbf{13.95} (15.02) & \textbf{6.16} (6.29) & \textbf{53.7} (53.0) & \textbf{42.4} (40.7) & \textbf{0.32} (0.30) & \textbf{0.61} (0.58) \\

4 & Back-Translation \cite{tan2019learning} & 9.81 (\textbf{9.62}) & 5.67 (\textbf{5.61}) & 54.8 (\textbf{54.9}) & \textbf{46.7} (46.6) & \textbf{0.43} (\textbf{0.43}) & 0.69 (\textbf{0.70}) \\

\bottomrule
\end{tabular}}
\captionsetup[table]{skip=10pt}
\caption{Comparison on the validation unseen split with and without the sub-instruction module. Values not in brackets are with sub-instructions, values in brackets are without sub-instructions.}
\label{tab:FiGraR2R_results}
\end{table*}
\noindent \textbf{Implementation details.} 
To obtain the word representations in each sub-instruction, the entire instruction is first passed to a unidirectional LSTM, then we implement chunking on the language hidden states to obtain the word representations of the selected sub-instructions. The ground-truth shifting signal at each time-step is dependent on the distance between the agent's current position and the end viewpoint of the selected sub-instruction. If the distance is smaller than or equal to 0.5 meters, the ground-truth shift signal ${s}_{t}$ will be $1$, and $0$ otherwise. For the Back-Translation model \cite{tan2019learning}, we only apply chunk shifting loss to the teacher-forcing imitation learning branch, so that the agent navigates on the ground-truth path and learns the chunk-shifting with less noise. We train all agents on a single NVIDIA Tesla K80 GPU, using the same hyperparameters as the baselines. 

\noindent \textbf{Evaluation metrics.} We follow the standard metrics that previous work employed for evaluating the agent's performance on the R2R dataset \cite{anderson2018vision}, which include Path Length (PL) of the agent's trajectory, average Navigation Error (NE) for the distance between agent's final position and the target, Oracle Success Rate (OSR) for the ratio of agents which the shortest distance between the target and the trajectory is within $3m$, Success Rate (SR) for the ratio of agents which the distance between agent's final position and the target is within $3m$, and Success Rate Weighted by Path Length (SPL). Furthermore, we also consider the normalized Dynamic Time Warping (nDTW) score \cite{magalhaes2019effective}, which is a metric that measure the overall performance of the agent with a focus on the similarity between the ground-truth and the actual trajectories.

\section{Results and Analysis}
\label{sec:results}

\begin{table*}[!ht]
\centering
\resizebox{0.94\textwidth}{!}{
\begin{tabular}{llc|cccc|cccc}
\toprule
\# & Model with sub-instructions & SR & TP & TN & FP & FN & Accuracy & Precision & Recall & F1-Score \\ \hline
1 & Seq2Seq \cite{anderson2018vision} & 22.9 & 608 & 36344 & 1602 & 4796 & 0.852 & 0.275 & 0.113 & 0.160 \\ 
2 & Speaker-Follower \cite{fried2018speaker} & 34.7 & 963 & 9966 & 452 & 4878 & 0.672 & 0.681 & 0.165 & 0.265 \\
3 & Self-Monitoring \cite{ma2019self} & 42.4 & 1130 & 10619 & 363 & 4686 & 0.699 & 0.757 & 0.194 & 0.309 \\ 
4 & Back-Translation \cite{tan2019learning} & 46.7 & 1256 & 8086 & 303 & 4765 & 0.648 & 0.806 & 0.209 & 0.331 \\
\bottomrule
\end{tabular}}
\caption{Statistics of the shifting signal on the unseeen validation set.}
\label{tab:confusion_matrix}
\end{table*}
We compare the performance of the four agents on the R2R unseen validation set. We also present the traceability of the navigation process resulting from our FGR2R data.

\subsection{Comparisons}
\noindent \textbf{Quantitative results}.
Table \ref{tab:FiGraR2R_results} shows the results of the four agents in unseen environments. The performance of the imitation learning agents (Row 1--3) with our sub-instruction attention module outperforms the base agents. In terms of the success rate, the Seq2Seq, Speaker-Follower and the Self-Monitoring agents achieve an absolute increase of 1.1\%, 4.9\% and 1.7\% respectively. The improvement is consistent in most of the other metrics, e.g. for the Self-Monitoring agent, its SPL improves from 0.30 to 0.32 and its nDTW score grows from 0.58 to 0.61. The overall improvement on Path Length and nDTW score for the first three agents indicates that using sub-instructions improves the agent's ability to navigate on the described path. As for the Back-Translation agent (Row 4), the performance with sub-instruction attention is very similar to the baseline, one possible reason could be that the introduction of sub-instruction shifting perturbs the learning of action during for the reinforcement learning scheme which the agent could deviate far from the ground-truth path.

Learning when the agent needs to read a new sub-instruction is a difficult task, the same viewpoint in a specific environment can be considered as a shifting point or not depending on the sub-instruction that the agent follows. 
In Table \ref{tab:confusion_matrix}, we show the confusion matrix of the shifting signals and we compute accuracy, precision, recall and F1-score to evaluate the performance of our proposed shifting module. Results show that all the agents have huge room for improvement for shifting, since the best F1-Score obtained is only 0.331. But we can see from the four agents that, as the success rate increases, the precision, recall and F1-score also improve. We propose to consider these results to be useful baselines for future methods that apply sub-instructions. Notice that agents visit a different number of viewpoints due to the maximum number of steps allowed, the use of panoramic action space and the ability to stop. In the case of Seq2Seq model, since the agent is not using a panoramic view, it performs many actions to change the camera orientation. 

\noindent \textbf{Qualitative performance}. 
We illustrate a qualitative example in Figure \ref{fig:path_comparison} to show how the sub-instruction module works in the agent. In the example, both the baseline model and the model with the sub-instruction module completes the task successfully. However, unlike the baseline model which fails to follow the instruction and stops within 3 meters of the target by chance, our model correctly identifies the completeness of each sub-instruction, guides the agent to walk on the described path and eventually stops right at the target position. We refer the readers to Supplementary Materials for visualization of more trajectories.


\begin{table}[!t]
\centering
\resizebox{0.48\textwidth}{!}{
\begin{tabular}{c|c|c|c|c|l}
\toprule
rank & $\overline{d}$ & $\overline{nDTW}$ & $f$ & $\overline{s}$ & \multicolumn{1}{c}{Representative sub-instruction} \\ \hline
1 & 2.22 & 0.72 & 7  & 2.8 & head down the stair \\
2 & 2.52 & 0.57 & 5  & 4.6 & wait near the first open door \\
3 & 2.58 & 0.73 & 8  & 2.5 & go into the bedroom \\
4 & 2.66 & 0.73 & 21 & 2.6 & exit the bedroom \\
5 & 2.77 & 0.65 & 10 & 2.1 & turn right at the entry\\

\vdots &\vdots &\vdots &\vdots &\vdots & \multicolumn{1}{c}{\vdots} \\
96  & 6.33   & 0.55 & 35    & 3.8   & stop behind the table at the far end \\
97  & 6.56   & 0.43 & 8     & 2.0   & walk past the sink, fridge, oven \\
98  & 6.86   & 0.46 & 11    & 3.2   & go through the wooden archway \\
99  & 6.88   & 0.51 & 20    & 3.0   & walk along the grass until you reach ... \\
100 & 7.36   & 0.52 & 38    & 3.4   & walk into the room which have a ... \\
\bottomrule
\end{tabular}}
\caption{Performance on different sub-instruction clusters in validation unseen split. $\overline{d}$, $f$ and $\overline{s}$ denote the mean distance, the frequency and the mean number of viewpoints of a cluster.}
\label{tab:cluster}
\vspace{-1em}
\end{table}
\subsection{Traceability}
With the FGR2R data, we reveal the navigation process of the agent working on specific sub-instructions. For each sub-instruction, we measure the similarity between the ground-truth path and the actual trajectory using nDTW as well as the distance between the end viewpoint of the sub-instruction and the predicted shift viewpoint. As a result, we can estimate the performance of the agent in each sub-task.

We cluster the sub-instructions into 100 clusters using complete-linkage hierarchical agglomerative clustering algorithm. Instead of using a standard metric of distance such as the Euclidean distance, we compute a similarity matrix of sub-instructions using the BLEU-4 metric. We experiment with the Self-Monitoring agent on validation unseen split and present a summary of the top five and the bottom five clusters ranked by the mean distance, as shown in Table~\ref{tab:cluster}.

We can see from the table that the clusters which the agent performs better consist of simple and direct sub-instructions which refer to a single action, such as ``\textit{head down the stair}" and ``\textit{exit the bedroom}". On the other hand, with sub-instructions that refer to specific objects such as ``\textit{walk past the sink, fridge, oven}'' or express an action which is conditioned on the completion of another action, such as ``\textit{walk along the grass until you reach ...}", the agent deviates far from the described path. Moreover, the ranking does not show a strong correlation with the frequency or the number of viewpoints of each sub-instruction. These results suggest that agent is incapable of understanding complex natural language instructions or ground to specific objects with a high accuracy.

\section{Conclusion}
\label{sec:conclusion}

In this paper we introduce a novel sub-instruction module and the Fine-Grained R2R Dataset to encourage the learning of correspondences between vision and language. The sub-instruction module enables the agent to attend to one particular sub-instruction at each time-step and decides whether the agent needs to proceed to the next sub-instruction. 
Our experiments show that by implementing the sub-instruction module in state-of-the-art agents, most of the agents are able to follow the given instruction more closely and achieve better performance. We also show that, with the sub-instruction annotations, the entire navigation trajectory is trackable. We believe that the idea of sub-instruction module and a sub-instruction annotated dataset can benefit future studies in the VLN task as well as other vision-and-language problems.

\bibliography{egbib}
\bibliographystyle{acl_natbib}

\appendix

\section{Appendices}
\label{sec:appendix}

\subsection{FGR2R Dataset}
\label{sec:appendix_A1}
\begin{figure*}[t!]
  \centering
  \includegraphics[width=0.73\textwidth]{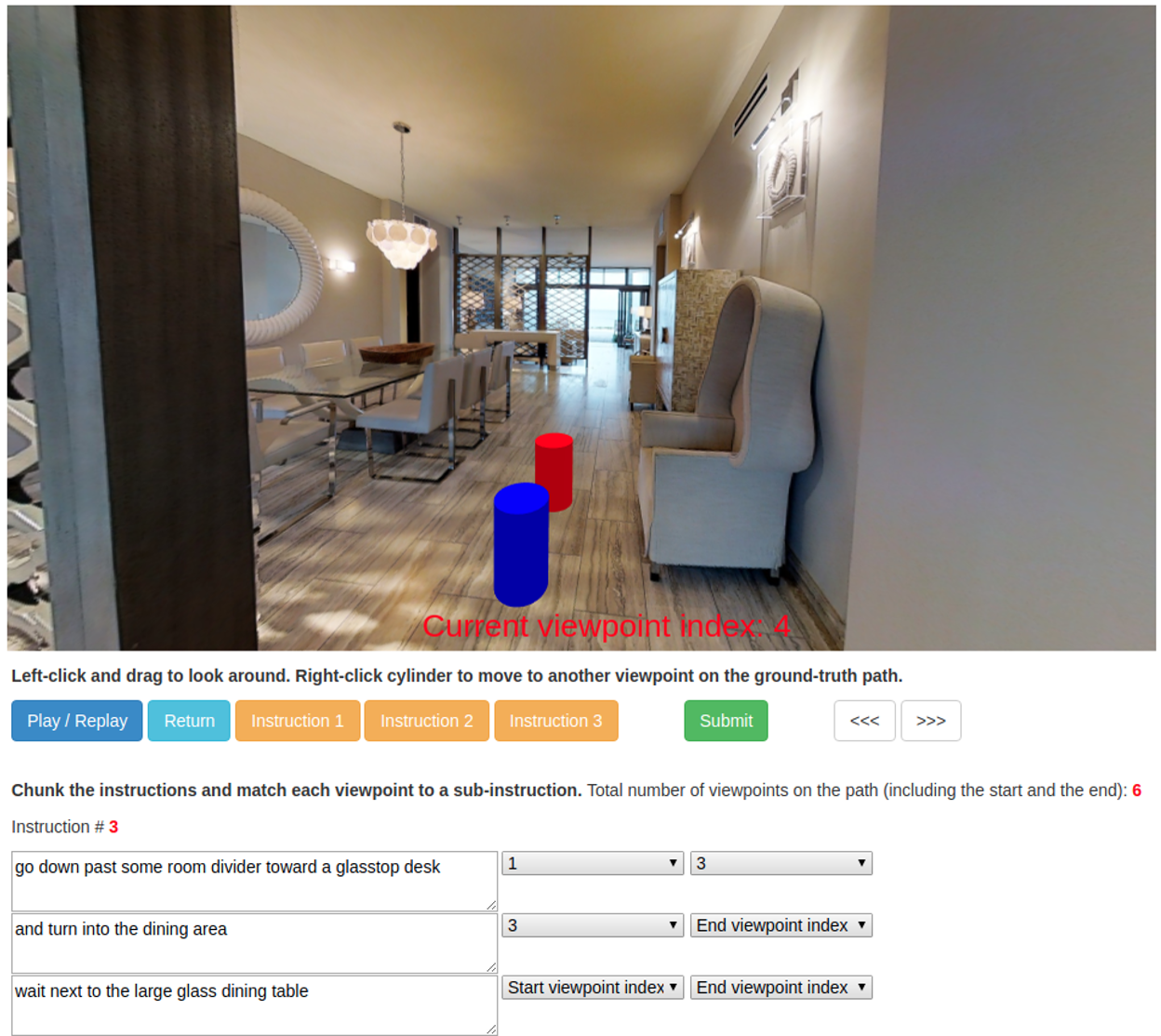}
  \captionof{figure}{The web interface for FGR2R data collection. The displayed photo of the environment is an interactive window, cylinders are the viewpoints on the ground-truth path. ``\textit{Play / Replay}'' shows an automatic run-through of the entire trajectory. ``\textit{Return}'' brings the agent back to the first viewpoint. ``\textit{Instruction \#}'' switches among the three instructions that described the same path. ``\textit{Submit}'' checks and submits the annotations.}
  \label{fig:interface}
  \vspace*{-5mm}
\end{figure*}

\noindent \textbf{Data collection.} We build a web interface to collect FGR2R data using Amazon Mechanical Turk (AMT), as shown in Figure~\ref{fig:interface}. In the interactive window, each viewpoint on the ground-truth path is highlighted with a large cylinder and an index of the viewpoint. Besides each sub-instruction, there is a drop-down list for assigning the start and end viewpoints of the corresponding sub-path. The annotators can click in the interactive window to freely move on the ground-truth path and freely rotate the camera to observe its surroundings. Before the start of labelling, we first ask the annotators to watch the automatic trajectory run-through to get familiar with the environment. Then, we ask them to partition the ground-truth path and assign a sub-instruction to those partitions. Once the labelling is completed, a function will automatically check if the annotation disobeys any rules (e.g., the start viewpoint of a sub-path should be the same as the end viewpoint of the previous sub-path) before approval for submission.

\begin{table*}[t!]
\centering
\resizebox{\textwidth}{!}{\begin{tabular}{llrrrrrr}
\toprule
& & \multicolumn{6}{c}{R4R Validation Unseen} \\
\cmidrule(r){3-8}
\# & Model & \multicolumn{1}{c}{PL} & \multicolumn{1}{c}{NE $\downarrow$} & \multicolumn{1}{c}{OSR $\uparrow$} & \multicolumn{1}{c}{SR $\uparrow$} & \multicolumn{1}{c}{SPL $\uparrow$} & \multicolumn{1}{c}{nDTW $\uparrow$} \\
\cmidrule(r){1-8}

1 & Seq2Seq \cite{anderson2018vision} & 9.40 (10.85) & 9.35 (\textbf{9.20}) & 32.8 (\textbf{35.5}) & 21.2 (\textbf{22.3}) & \textbf{0.11} (\textbf{0.11}) & 0.42 (\textbf{0.43}) \\

2 & Speaker-Follower \cite{fried2018speaker} & 26.64 (25.68) & 8.46 (\textbf{8.09}) & \textbf{42.1} (40.7) & 26.4 (\textbf{27.4}) & 0.12 (\textbf{0.13}) & \textbf{0.41} (\textbf{0.41}) \\

3 & Self-Monitoring \cite{ma2019self} & 28.01 (23.41) & \textbf{8.07} (8.46) & \textbf{46.2} (40.6) & \textbf{27.4} (25.8) & \textbf{0.10} (0.09) & \textbf{0.42} (0.41) \\

4 & Back-Translation \cite{tan2019learning} & 7.78 (39.66) & 9.33 (\textbf{7.90}) & 38.1 (\textbf{53.5}) & 21.5 (\textbf{31.2}) & \textbf{0.17} (0.14) & \textbf{0.48} (0.39) \\

\bottomrule
\end{tabular}}
\captionsetup[table]{skip=10pt}
\caption{Comparison on the R4R validation unseen split with and without the sub-instruction module. Values not in brackets are with sub-instructions, values in brackets are without sub-instructions.}
\label{tab:FiGraR4R_results}
\end{table*}
\noindent \textbf{Annotator qualification.} To ensure the quality of the annotation returned by the annotators, we annotated a subset of 300 samples as ground-truths and we exam each annotator with 15 ground-truth samples before approval for labelling. In total, there are 126 participants. We reject workers with a low agreement to the ground-truth. The qualification process leaves us 58 qualified annotators to complete the annotation task.

\begin{figure}
  \centering
  \includegraphics[width=0.48\textwidth]{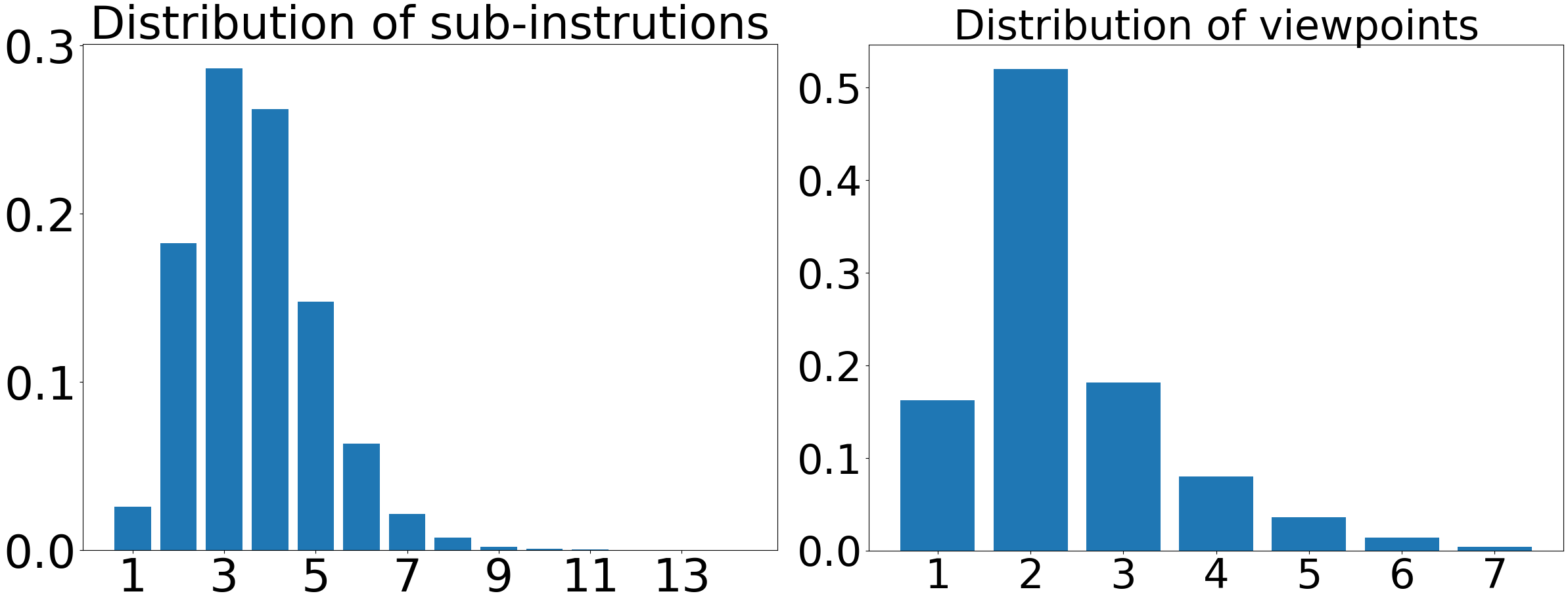}
  \captionof{figure}{Distribution of sub-instructions in an instruction and distribution of viewpoints for a sub-instruction in the FGR2R dataset.}
  \label{fig:distributions}
  \vspace*{-5mm}
\end{figure}

\noindent \textbf{Dataset statistics.} Apart from the FGR2R statistics mentioned in the paper, we present the distribution of sub-instructions in an instruction and the distribution of viewpoints for a sub-instruction in Figure \ref{fig:distributions}. As we can see, most of the instructions are broken down into more than one sub-instruction and the frequency of more than seven sub-instructions is very low. Also, notice that about 15\% of the sub-instructions are paired with only one viewpoint, as a result of the sub-instructions that only refer to camera rotation such as ``\textit{rotate slightly to the left}'' or stopping command such as ``\textit{wait by the sink}''.

\noindent \textbf{Training with FGR2R.} During training, consider that more coherent motion could be beneficial for the agent to learn the textual-visual correspondence. We combine the sub-instructions which are only paired with one viewpoint to the next sub-instruction (and combine with the previous sub-instruction if it is the last one). The sub-instructions in validation sets remain in their original format so that the ground-truth trajectories are kept unknown. In this work, we only enable the sub-instruction module with a single step uni-directional shifting, which agrees with the observation that instructions and trajectory in the R2R dataset are monotonically aligned. However, different rules could be designed. For example, one can allow the agent to shift for more than one step or enable the agent to read the previous sub-instructions once it backtracks to the visited viewpoint. Our proposed FGR2R make all these research directions possible. We leave these ideas to future research.

\subsection{Extension to Fine-Grained R4R} 

\noindent \textbf{R2R to R4R} The R4R dataset is created by concatenating two trajectories in R2R, which the first path ends within three meters from the start of the second path \cite{jain2019stay}. We enrich the R4R data with sub-instructions annotations by joining two sequences of sub-instructions corresponding to the two trajectories. However, for some trajectories in R4R, there exist several additional viewpoints for connecting the two paths, which has no sub-instruction annotation. Therefore, we assign those additional viewpoints to the first sub-instruction of the second path.

\noindent \textbf{Evaluation} We further experimented the four agents on the R4R dataset, with and without sub-instruction modules. As shown in Table~\ref{tab:FiGraR4R_results}, the performance of the first three agents are very similar. For agents with sub-instruction modules, the SR of Seq2Seq and Speaker-Follower are slightly lower, whereas the SR of Self-Monitoring agent is 1.6\% higher. As for the Back-Translation, the agent experiences a large OSR and SR drop after applying sub-instructions, but the SPL and nDTW are increased by 3\% and 9\%. This result indicates that although the agent with sub-instruction modules has a lower chance to reach the target (stop within 3m), it follows the instruction much better.

However, we argue that a large performance gain has not been obtained in R4R mainly for two reasons: (1) The additional viewpoints created for linking the two trajectories have no corresponding sub-instructions. Hence, agents trained to follow each sub-instructions strictly have no guidance for those steps. (2) The last sub-instruction of the first trajectory is very confusing to the agent, as it usually refers to the $STOP$ action, but the navigation does not end. This prevents the agent from learning a good stopping policy since the ground-truth action requires the agent to keep moving.

In conclusion, we believe that it is inappropriate to apply FGR2R data directly for FGR4R task. To obtain FGR4R data, our suggestion is to remove the final sub-instruction about the $STOP$ action from the first trajectory, and use a Speaker module \cite{fried2018speaker} to generate a new sub-instruction for the additional viewpoints for linking the two trajectories. We will leave this idea for future work.

\subsection{Visualization of Navigation}
We visualize the navigation trajectories of the Self-Monitoring agent with and without our proposed sub-instruction module in the following pages.





\clearpage

\begin{figure*}
  \centering
  \includegraphics[width=0.93\textwidth]{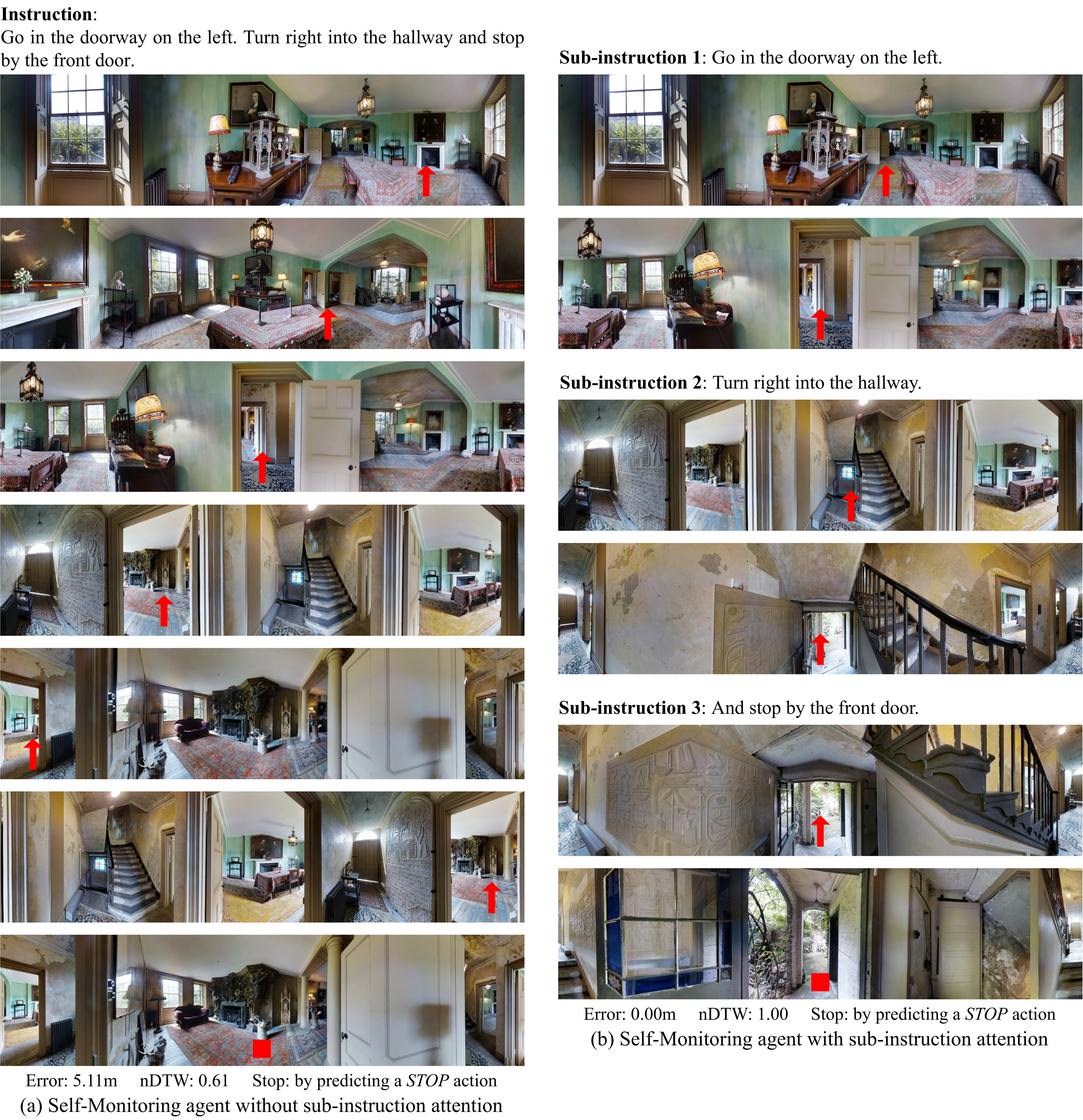}
  \captionof{figure}{A positive example of sub-instruction aware navigation. Without sub-instruction module, the agent wanders between rooms and decides to stop at a wrong location. With sub-instruction module, the agent successfully leaves the room, finds the way to the target and stops at the right location.}
  \label{fig:supmat_goodpath_5}
\end{figure*}

\begin{figure*}
  \centering
  \includegraphics[width=0.93\textwidth]{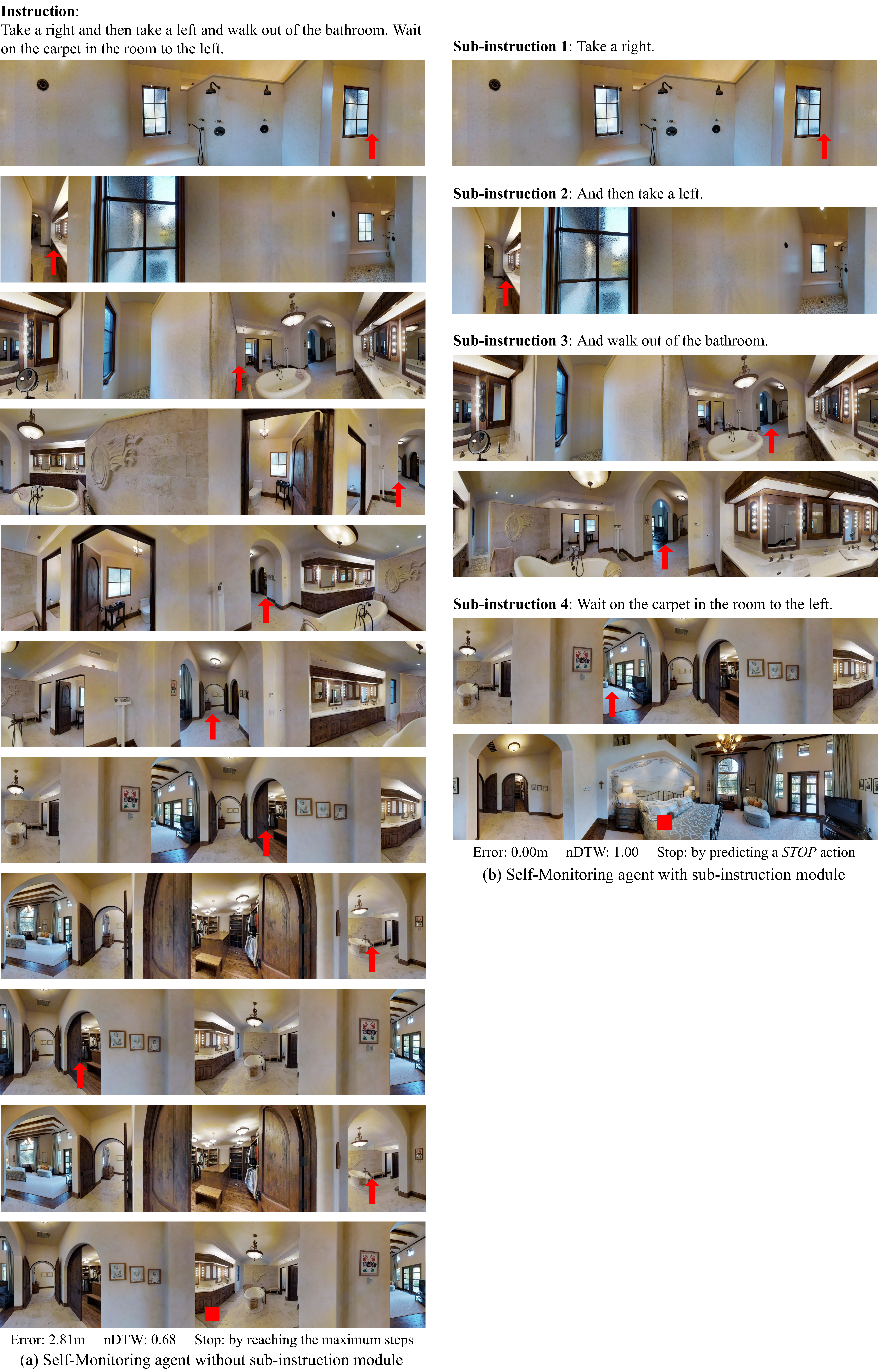}
  \caption{A positive example of sub-instruction aware navigation. Without sub-instruction module, the agent fails to follow the instruction and stops next to the target by chance. With sub-instruction module, the agent navigates on the described path and eventually stops right at the target location.}
  \label{fig:supmat_goodpath_2}
\end{figure*}

\begin{figure*}
  \centering
  \includegraphics[width=0.93\textwidth]{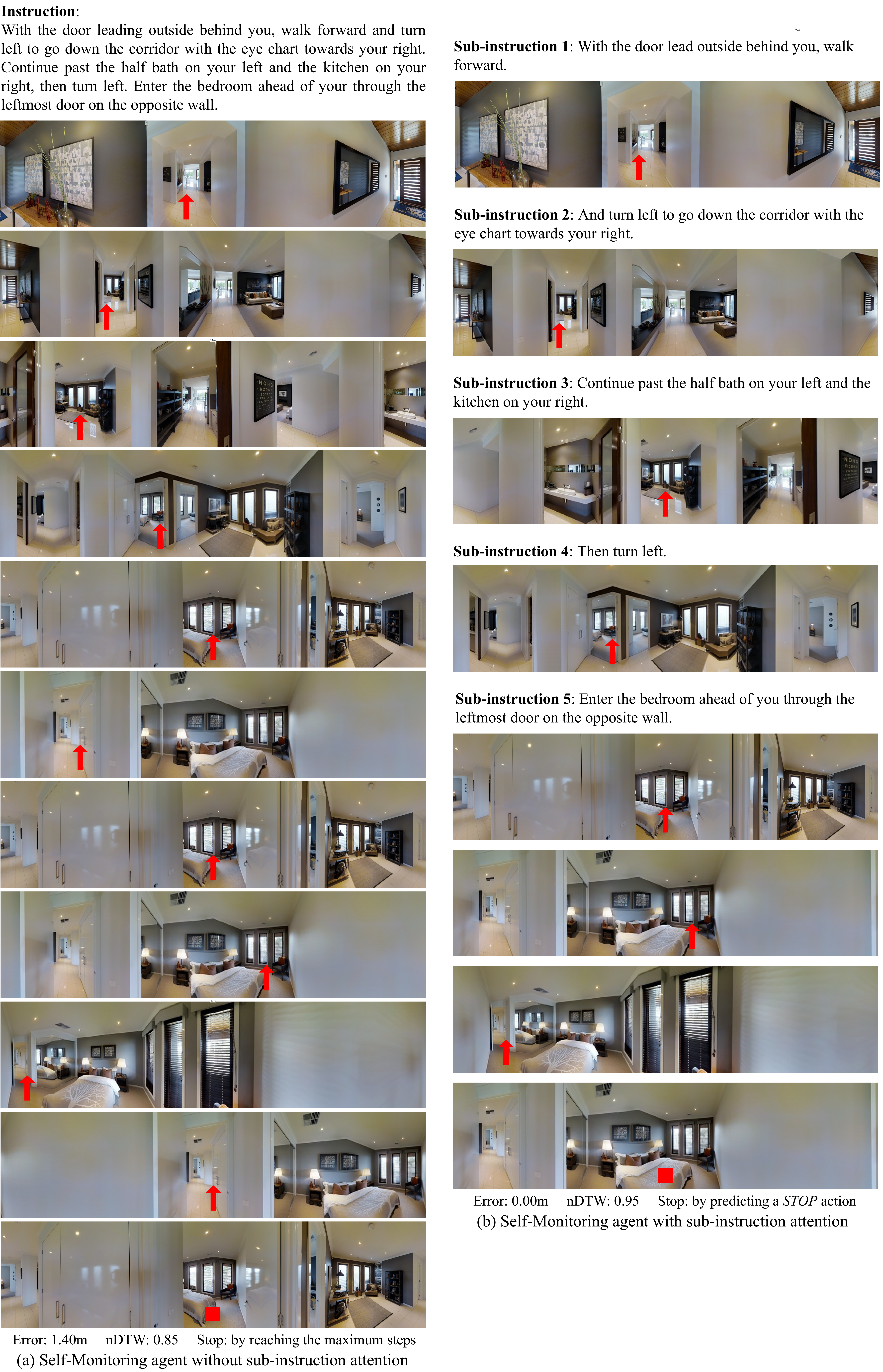}
  \caption{A positive example of sub-instruction aware navigation. Without sub-instruction module, the agent loops around the target and doesn't know how to stop. With sub-instruction module, the agent falls into the same loop but quickly escapes from it and stops at the correct location.}
  \label{fig:supmat_goodpath_6}
\end{figure*}

\begin{figure*}
  \centering
  \includegraphics[width=0.93\textwidth]{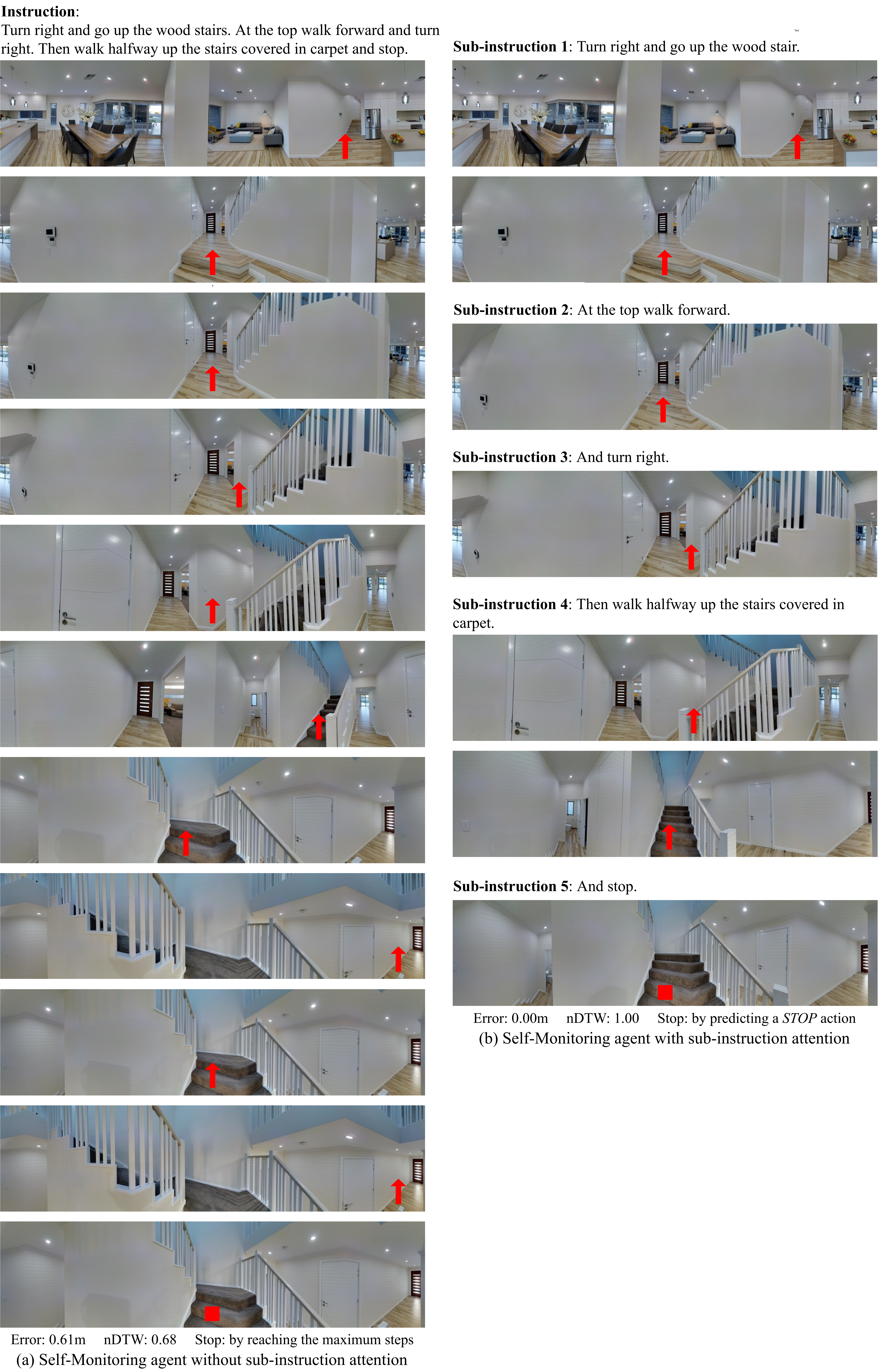}
  \caption{A positive example of sub-instruction aware navigation. Without sub-instruction module, the agent loops around the target and doesn't know how to stop. With sub-instruction module, the agent navigates on the described path and eventually stops right at the target location.}
  \label{fig:supmat_goodpath_3}
\end{figure*}


\begin{figure*}
  \centering
  \includegraphics[width=0.93\textwidth]{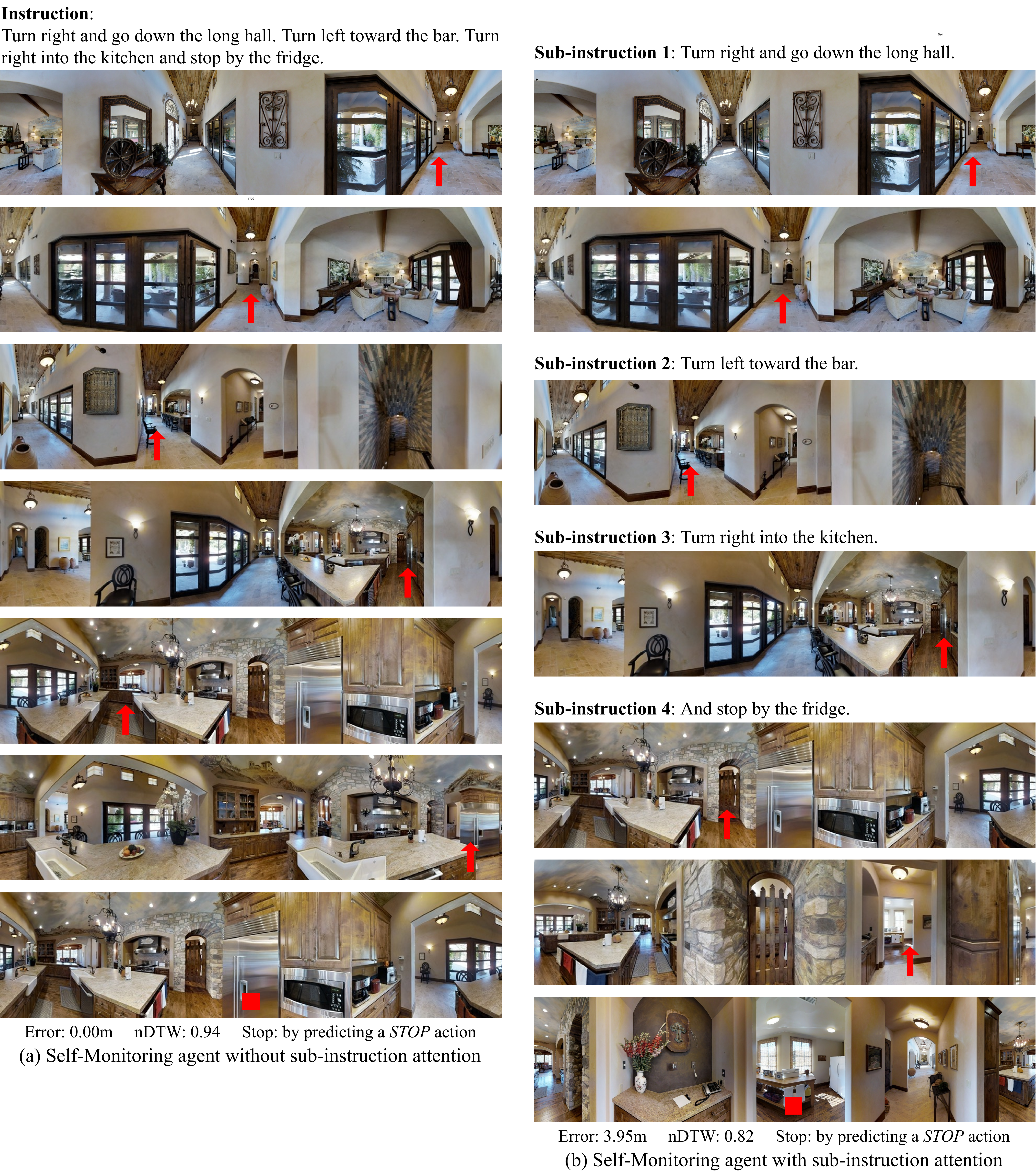}
  \caption{A negative example of sub-instruction aware navigation. Without sub-instruction module, the agent completes the navigation task without making any mistake. With sub-instruction module, although the agent performs sub-instruction shifting perfectly, it overlooks the target object and walks away from the target, eventually decides to stop at a wrong location.}
  \label{fig:supmat_goodpath_7}
\end{figure*}

\end{document}